\documentclass[11pt,a4paper]{article}
\usepackage[hyperref]{emnlp2020}
\usepackage{times}
\usepackage{float}
\usepackage{appendix}
\usepackage{latexsym}

\usepackage{microtype}
\aclfinalcopy 

\title{Detecting Gender Bias in Course Evaluations}

\author{Sarah Lindau \\
Chalmers University of Technology \\
  \texttt{lsarah@student.chalmers.se} \\\And
  Linnea Nilsson \\
  Chalmers University of Technology \\
  \texttt{linsvens@student.chalmers.se} \\}

\date{\today}

\begin{document}
\maketitle
\begin{abstract}
A master thesis studying gender bias in course evaluations through the lense of machine learning and nlp. We use different methods to examine and explore the data and find differences in what students write about courses depending on gender of the examiner. Data from English and Swedish courses are evaluated and compared, in order to capture more nuance in the gender bias that might be found. Here we present the results from the work so far, but this is an ongoing project and there is more work to do. 
\end{abstract}

\section{Introduction}
This project is financed by the Chalmers initiative  \href{https://www.chalmers.se/en/about-chalmers/Chalmers-for-a-sustainable-future/initiatives-for-gender-equality/gender-initiative-for-excellence/Pages/default.aspx}{GENIE (Gender Initiative for Excellence)},
and is a part of the \href{http://www.cse.chalmers.se/~peb/genie-project.html}{GENIE project: Analysis of gender bias}, a project  that aims to use Natural Language Processing techniques to investigate gender bias in texts of different genres linked to Chalmers University of Technology and Gothenburg University. As our contribution to this project, we are currently writing a master thesis in which we explore course evaluations written by students at Chalmers University of Technology in order to evaluate gender bias in an educational context.
\begin{figure}[h]
    \centering
    \includegraphics[width=0.45\textwidth]{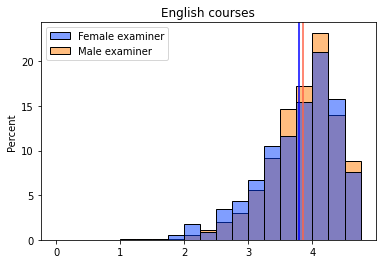}
    \includegraphics[width=0.45\textwidth]{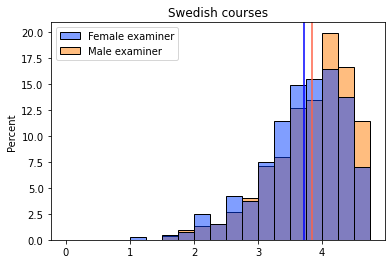}
    \caption{Overall impression for courses with male and female examiners, separated by teaching language.}
    \label{fig:gender_eng_swe_comparison}
\end{figure}

While the project consists of numerous research questions and experiments, this submission will focus on the experiment where we try to find out if there's any lingustic differences in the comments provided by students in course evaluations depending on the gender of the examiner. In order to find such differences we train classifiers with the texts from the course evaluations and use them in order to predict the gender of the examiner.

\section{Background}
When working with the course evaluations we found that the overall impression students have of courses was lower for courses with female examiners. The overall impression is measured as a 1-5 score in the course evaluations, as well as a free text answer. We found that the mean score for courses with a female examiner were lower than the courses with a male examiner. The mean scores were 3.747 for female lead courses and 3.848 for male lead course, which result in a difference of 0.101 points lower for female examiners. The differences in mean scores considering the gender of the examiner were greater for courses held in Swedish compared to courses held in English. To visualize these differences we refer to figure \ref{fig:gender_eng_swe_comparison}. These results could indicate a bias against female examiners and therefore it would be a fair guess that some differences in the language style and wording in the evaluations would indeed differ depending on the gender of the examiner.  

Previous research suggest that there is gender bias against female teachers. A 2018 dutch study clearly shows the existence of gender bias against female teachers in university. \citep{teacherseval} The study examined almost 20000 teacher evaluations finding that female instructors systematically receieved lower scores. In order to rule out the cause being difference in subject or difficulty of the specific course being evaluated the students were randomly assigned either male or female instructors within the same course and the evaluations were collected before the final exam and grading of the students. To rule out that women simply are inferior teachers the study also includes data of the students grades and self estimated study hours and concludes that there are no significant difference in the grades or the study effort of the students based on the gender of the instructor.

When working with gender it's important to note that sex and gender are not the same thing, while they are related. \citep{klauscomputer} A persons sex is typically either determined as male or female and is determined by biological factors such as their chromosomes. Gender is a wider term and is rather concerned with cultural and behavioral aspects related to sex. \cite{gendersex}

\section{Data}
The data for this project consists of course evaluations from 9165 Chalmers courses from 2013 to 2021, a file with anonymized student grades, a list of examiners per course and finally Swedish name statistics from Statistiska CentralByrån. The main dataset is the course evaluations, this is also where all text data is located. The other data is used to provide interesting metadata for the project. The course evaluations were originally structured as one file per course, most in an excel format and some in a pdf-format. The pdf files were excluded in order to simplify the processing of the data. The dataset contains course evaluations in Swedish and English depending on the teaching langauge of the course.

When choosing what data to analyze, we had to select courses based on some criteria. First, a course that was to be selected needed to have all the interesting data. That is, we needed to have evaluations, student grades and where able to give a conclusive prediction of the examiner gender using the Swedish name statistics. In order to avoid basing to much of our results on the outliers, we also excluded courses that had fewer than 25 students and less than 10\% female students. After the final selection, we ended up with 4535 courses. All data available about these courses was put in a json file that all work was based on. The data for each course consists of the evaluation questions, the student answers, as well as metadata including the course name, study period, year, course code, the number of students of each gender that received each grade, gender of the examiner and the course language. 

\section{Methods}
In order to work with the text data, we used a bag-of-words model, using a count vectorizer from sklearn on all the text comments. One trained on the English data and one on the Swedish data. The student answers needed to be anonymized as they might contain the examiner name which could affect tasks such as examiner gender prediction. For the same reason, we chose to remove all gendered pronouns, such as his, her, hans or hennes. This was done by creating a list of all words to remove and then using it as stopwords input to the count vectorizer. We chose to create several different versions of the dataset in order to see the effect that some preprocessing steps could have further down the line. Those steps were undersampling and including part of speech tags. For the undersampling, we created our own method using the python random.sample method and set the random seed to 1 for reproducability. Using the random.sample method, we simply selected as many samples in the majority class as we have samples in the minority class. The part of speech tags were generated using the nlp library Spacy. The word lemmas and the POS tags where put in a list that was then fed through the count vectorizer. 

Training and testing data was separated randomly using 20\% of the data as test data. However, for the continuation of the work and to get more robust results we may instead use something along the lines of cross validation. 

The examiner gender prediction was performed using two different classifiers, a logistic regression classifier and a random forest classfier. Both models from sklearn. A random forest is a relatively common ensemble model that is based on decision trees. It works by combining the results from several decision trees. These are trained individually on randomly selected parts of the training data, aswell as using randomly selected features to split the data. \citep{WhatIBM1} Logistic regression is another type of classification algorthm. Simply explained, this model bases predictions on a probability score that makes the user able to determine how confident the model is of the classification.\citep{logreg}

In order to evaluate the results from the classifiers, we wanted to see what features had been the most important for them to predict the two classes (male or female). 

When working with the examiner gender in this project we have given them labels "male" and "female" which would typically be associated with sex rather than gender. However, what we are actually investigating is the students's perception of their examiner, which would relate to their gender. Since we don't have access to the examiners self-reported gender identities, we have used their first names together with Swedish name statistics to give an estimation of their gender.

\section{Results and Discussion}
The main purpose of the examiner gender classification is not really to solve the classification task of predicting the examiner gender from the text comments. Instead, we are examining whether or not the classfiers are able to solve the task sufficiently, which would indicate that there is some difference to how students write about their courses and examiners based on their gender. For this reason, the main results are focused on what features that the classifiers have deemed important rather than the classification accuracy or f1-score. Still, we need to note that a better classifier has found a stronger way to predict and thus should give us clearer indications as to what features are important, so we still need to look at the more traditional evaluation methods.

In order to evaluate our classifiers, we need a baseline for comparison. Since our dataset is unbalanced with a lot more male examiners we need to take that into account. So for our baseline we only use the most common label (male).

\begin{center}
\begin{table}[H]
\begin{tabular}{ |c |c c|}
\hline
\multicolumn{3}{|c|}{Baseline} \\
\hline
  Result & Eng & Swe \\
  \hline
  Accuracy & 0.80000 & 0.84069\\ 
  Precision & 0.80000 & 0.84069\\  
  Recall & 1.00000 & 1.00000\\
  ROC AUC & 0.50000 & 0.50000\\
  \hline
\end{tabular}
\caption{The results from the baseline model.}
\label{tab:baseline}
\end{table}
\end{center}
When evaluating the logistic regression model on Swedish data (see table \ref{tab:logregswe}), we found that the accuracy and recall scores were significantly worse for the undersampled data. This was not entirely surprising due to the unbalanced nature of the testing data compared to the now balanced training data. When training the models on undersampled data we also used less training samples, which could effect the robustness of the model. 
\begin{center}
\begin{table}[H]
\begin{tabular}{ |c |c c c |}
\hline
\multicolumn{4}{|c|}{Logistic Regression Swedish} \\
\hline
  Result & text & POS & US \\
  \hline
  Accuracy &  0.76967 & 0.79079 & 0.61228\\ 
  Precision & 0.85177 & 0.87302 & 0.87107\\  
  Recall & 0.87900 & 0.87900  & 0.63242\\
  ROC AUC & 0.53588 & 0.60215 & 0.56922\\
  train samples & 2080 & 2080 & 654\\
  \hline
\end{tabular}
    \caption{The results from training the logistic regression model on Swedish data. Text is trained on all the text data, without part-of-speech tags, for POS the part-of-speech tags are added and finally US is the undersampled text data without POS tags. }
    \label{tab:logregswe}
\end{table}
\end{center}

\begin{center}
\begin{table}[H]
\begin{tabular}{ |c |c c c |}
\hline
\multicolumn{4}{|c|}{Logistic Regression English} \\
\hline
  Result & text & POS & US \\
  \hline
  Accuracy & 0.76104 & 0.74545 & 0.62338\\ 
  Precision &  0.85762 & 0.84768 & 0.88626\\  
  Recall & 0.84091 & 0.83117 & 0.60714\\
  ROC AUC & 0.64123 & 0.61688 & 0.64773\\
  train samples & 1538 & 1538 & 542\\
  \hline
\end{tabular}
    \caption{The results from training the logistic regression model on English data. Text is trained on all the text data, without part-of-speech tags, for POS the part-of-speech tags are added and finally US is the undersampled text data without POS tags. }
    \label{tab:logregeng}
\end{table}
\end{center}
It is interesting to note that this difference was smaller for the English data in our experiments. Although the English model performed slightly worse overall, as can be seen in table \ref{tab:logregeng}. When using the undersampled data, we would hope that the models would be able to correctly classify more of the female samples as these are not "drowned out" by the male. In the random forest model we have seen similar results, but it seems to perform better overall, as can be seen in tables \ref{tab:randforswe} and \ref{tab:randforeng}.

\begin{center}
\begin{table}[H]
\begin{tabular}{ |c |c c c |}
\hline
\multicolumn{4}{|c|}{Random Forest Swedish} \\
\hline
  Result & text & POS & US \\
  \hline
  Accuracy & 0.84069 & 0.84069 & 0.65835\\ 
  Precision & 0.84069 & 0.84069 & 0.90373\\  
  Recall & 1.00000 & 1.00000 & 0.66438\\
  ROC AUC & 0.50000 & 0.50000 & 0.64544\\
  \hline
\end{tabular}
    \caption{The results from training the random forest model on Swedish data. Text is trained on all the text data, without part-of-speech tags, for POS the part-of-speech tags are added and finally US is the undersampled text data without POS tags. }
    \label{tab:randforswe}
\end{table}
\end{center}

\begin{center}
\begin{table}[H]
\begin{tabular}{|c |c c c |}
\hline
\multicolumn{4}{|c|}{Random Forest English} \\
\hline
  Result & text & POS & US \\
  \hline
  Accuracy & 0.80000 & 0.79740 & 0.64156\\ 
  Precision & 0.80000 & 0.79948 & 0.89352\\  
  Recall & 1.00000 & 0.99675 & 0.62662\\
  ROC AUC & 0.50000 & 0.49838 & 0.66396\\
  train samples & 2080 & 2080 & 654\\
  \hline
\end{tabular}
    \caption{The results from training the random forest model on English data. Text is trained on all the text data, without part-of-speech tags, for POS the part-of-speech tags are added and finally US is the undersampled text data without POS tags. }
    \label{tab:randforeng}
\end{table}
\end{center}

The main reason for training the classifiers was to see  if we could find any differences in what features they found important for classifying a sample as male or female. For this purpose lists of the ten most important features where produced for each classifier instance trained on a dataset. These lists can be found in the appendix. While the evaluation of these lists may become a bit subjective, it is still interesting to compare these lists. 

For the Swedish data, we see that many words are related to school and schoolwork both for predicting male and female, such as "kurslitteraturen" (the course literature) and "materialen" (the materials). There is some difference to the words that are used, which indicates that there is some difference to what students write about the courses. However, there is no clear and easily distinguishable pattern to it.

For the English data, the differences seem a bit clearer. When predicting a female examiner, typically "soft" words, such as "open", "feels" and "writing" are important and can be found in table \ref{tab:words_logreg_eng_F}. This can be compared to the words used to predict male examiners in table \ref{tab:words_logreg_eng_M}, where we find words such as "harder", "clearer" and "process". In conclusion, our results thus far seem to indicate some differences to what students write about courses depending on the gender of the examiner.

\section{Future Work} 
This master thesis project is not yet finished and there are still things that should be further examined and investigated. The analysis of what features are important to the examiner gender classifiers needs to be performed on the random forest classifiers aswell. This analysis also needs to deepen, to find meaning to what differences there are and what that indicates.

To further explore the differences about how students write, we plan to perform a similar classification of author gender. This will, however, be limited by the anonymity of the students, but can be approximated by gender distribution in the class. 

It would be interesting to explore the texts in the course evaluations using word embeddings to see if we could capture more differences to the language that is used. These result could potentially be used to detect wether or not there is any gender bias present in the data.

In conclusion, to really capture any gender bias in the course evaluations it needs to be further explored and examined within the scope of the project. For future projects, it would be interesting to see how these results translate, both geographically as well as for different fields of education.

\bibliography{emnlp2020.bib}

\begin{thebibliography}{5}
\expandafter\ifx\csname natexlab\endcsname\relax\def\natexlab#1{#1}\fi

\bibitem[{Bandgar(2021)}]{logreg}
Swapnil Bandgar. 2021.
\newblock \href {https://medium.com/analytics-vidhya/logistic-regression-c5a6c047363e} {{Logistic Regression.}}

\bibitem[{Education(2020)}]{WhatIBM1}
IBM~Cloud Education. 2020.
\newblock \href {https://www.ibm.com/cloud/learn/random-forest} {{What is Random Forest? | IBM}}.

\bibitem[{Mengel et~al.(2018)Mengel, Sauermann, and Zölitz}]{teacherseval}
Friederike Mengel, Jan Sauermann, and Ulf Zölitz. 2018.
\newblock \href {https://doi.org/10.1093/jeea/jvx057} {{Gender Bias in Teaching Evaluations}}.
\newblock \emph{Journal of the European Economic Association}, 17(2):535--566.

\bibitem[{Scheuerman et~al.(2019)Scheuerman, Paul, and Brubaker}]{klauscomputer}
Morgan~Klaus Scheuerman, Jacob~M. Paul, and Jed~R. Brubaker. 2019.
\newblock \href {https://doi.org/10.1145/3359246} {How computers see gender: An evaluation of gender classification in commercial facial analysis services}.
\newblock \emph{Proc. ACM Hum.-Comput. Interact.}, 3(CSCW).

\bibitem[{Torgrimson and Minson(2005)}]{gendersex}
Britta~N. Torgrimson and Christopher~T. Minson. 2005.
\newblock \href {https://doi.org/10.1152/japplphysiol.00376.2005} {Sex and gender: what is the difference?}
\newblock \emph{Journal of Applied Physiology}, 99(3):785--787.
\newblock PMID: 16103514.

\end{thebibliography}
\bibliographystyle{acl_natbib}

\newpage
\appendix
\appendixpage
\addappheadtotoc

\section{Important features}
Here we have gathered the ten most important features to predict the examiner as male and female for the logistic regression classifiers that are trained on English and Swedish data. We include the results from all three versions of the dataset: undersampled, including the part of speech tags and just all text data.

\begin{center}
\begin{table}[H]
\begin{tabular}{ |c c c|}
\hline
\multicolumn{3}{|c|}{Important features} \\
\hline
  Text & US & POS\\
  \hline
  ännu & hp & ännu\\
  senare & tar & etc\\
  materialet & ännu & lämna\\
  arbeta & chalmers & förstod\\
  skönt & låg & tempo\\
  flesta & flesta & dra\\
  etc & följa & särskilt\\
  inlämningsuppgifter & laborationen & extrem\\
  varandra & tiden & välja\\
  bästa & mesta & lärande\\
  \hline
  
\end{tabular}
    \caption{The ten most important features for predicting a sample as female for the logistic regression model trained and evaluated on Swedish data.}
    \label{tab:words_logreg_swe_F}
\end{table}
\end{center}

\begin{center}
\begin{table}[H]
\begin{tabular}{ |c c c|}
\hline
\multicolumn{3}{|c|}{Important features} \\
\hline
  Text & US & POS\\
  stressful & based & structured\\
  works & stressful & stressful\\
  except & important & another\\
  three & writing & correct\\
  open & works & base\\
  based & test & consider\\
  feels & examples & until\\
  giving & when & reason\\
  change & period & leave\\
  background & small & opportunity\\
  \hline
  \hline
\end{tabular}
    \caption{The ten most important features for predicting a sample as female for the logistic regression model trained and evaluated on English data.}
    \label{tab:words_logreg_eng_F}
\end{table}
\end{center}

\begin{center}
\begin{table}[H]
\begin{tabular}{ |c c c|}
\hline
\multicolumn{3}{|c|}{Important features} \\
\hline
  Text & US & POS\\
    computer & tutorials& professor\\
  harder & way & computer\\
  tutorials & might & okay\\
  code & det & page\\
  perfect & over & code\\
  those & problem & introduce\\
  process & page & form\\
  such & needs & perfect\\
  learnt & exams & overall \\
  clearer & computer& tool\\
  \hline
  \hline
\end{tabular}
    \caption{The ten most important features for predicting a sample as male for the logistic regression model trained and evaluated on English data.}
    \label{tab:words_logreg_eng_M}
\end{table}
\end{center}

\begin{center}
\begin{table}[H]
\begin{tabular}{ |c c c|}
\hline
\multicolumn{3}{|c|}{Important features} \\
\hline
  Text & US & POS \\
  \hline
     nytt & gått & kvar\\
  förväntades & rolig & handledarna\\
  givande & möjlighet & faktisk\\
  hjälpte & där & pga\\
  rolig & bort & praktisk\\
  tänka & dessutom & nivån\\
 däremot & heller & tentorna\\
  klart & ger & ex\\
  ger& dom & skapa\\
  kurslitteraturen&la & lärorik\\
  \hline
\end{tabular}
    \caption{The ten most important features for predicting a sample as male for the logistic regression model trained and evaluated on Swedish data.}
    \label{tab:words_logreg_swe_M}
\end{table}
\end{center}

\end{document}